\documentclass[conference]{IEEEtran}
\IEEEoverridecommandlockouts
\usepackage{cite}
\usepackage{amsmath,amssymb,amsfonts}
\usepackage{graphicx}
\usepackage{textcomp}
\usepackage{xcolor}
\usepackage{caption}
\usepackage{newtxtext,newtxmath}
\usepackage{multicol} 
\usepackage{subcaption}
\usepackage{amsmath}
\usepackage{algorithm2e}

\def\BibTeX{{\rm B\kern-.05em{\sc i\kern-.025em b}\kern-.08em
    T\kern-.1667em\lower.7ex\hbox{E}\kern-.125emX}}
\begin{document}

\title{Developing, Analyzing, and Evaluating Vehicular Lane Keeping Algorithms Under Dynamic Lighting and Weather Conditions Using Electric Vehicles}

\author{
  \IEEEauthorblockN{Michael Khalfin\IEEEauthorrefmark{1},
                    Jack Volgren\IEEEauthorrefmark{2},
                    Matthew Jones\IEEEauthorrefmark{3},
                    Luke LeGoullon\IEEEauthorrefmark{4},
                    Joshua Siegel\IEEEauthorrefmark{5}, and
                    Chan-Jin Chung\IEEEauthorrefmark{6}}
  \IEEEauthorblockA{\IEEEauthorrefmark{1}Department of Computational Applied Mathematics, Rice University, mlk15@rice.edu}
  \IEEEauthorblockA{\IEEEauthorrefmark{2}Department of Computer Science and Engineering, Pennsylvania State University, jqv5334@psu.edu}
  \IEEEauthorblockA{\IEEEauthorrefmark{3}Department of Mathematics and Computer Science, Willamette University, mpjones@willamette.edu}
  \IEEEauthorblockA{\IEEEauthorrefmark{4}Division of Computer Science and Engineering, Louisiana State University, llegou1@lsu.edu}
  \IEEEauthorblockA{\IEEEauthorrefmark{5}Department of Computer Science and Engineering, Michigan State University, jsiegel@msu.edu}
  \IEEEauthorblockA{\IEEEauthorrefmark{6}Department of Mathematics and Computer Science, Lawrence Technological University, cchung@ltu.edu}
}

\maketitle

\begin{abstract}
  Self-driving vehicles have the potential to reduce accidents and fatalities on the road. Many production vehicles already come equipped with basic self-driving capabilities, but have trouble following lanes in adverse lighting and weather conditions. Therefore, we develop, analyze, and evaluate two vehicular lane-keeping algorithms under dynamic weather conditions using a combined deep learning- and hand-crafted approach and an end-to-end deep learning approach. We use image segmentation- and linear-regression based deep learning to drive the vehicle toward the center of the lane, measuring the amount of laps completed, average speed, and average steering error per lap. Our hybrid model completes more laps than our end-to-end deep learning model. In the future, we are interested in combining our algorithms to form one cohesive approach to lane-following.
\end{abstract}


\section{Introduction}

An estimated 9,330 people from the United States died in motor vehicle accidents in the first quarter of 2023 \cite{b1}. Additionally, for seven consecutive quarters since 2020, fatalities increased, likely due to the Covid-19 Pandemic.

Self-driving vehicles have the potential to reduce fatalities due to human error. People believe that automated vehicles should be four to five times safer than human-driven vehicles, or the traffic risk for automated vehicles should be two orders of magnitude lower than human-driven vehicles \cite{b8}. Self driving must become more consistent and reliable in order for this standard to be met.

\begin{figure}[tb]
  \centering
  \includegraphics[width=\columnwidth]{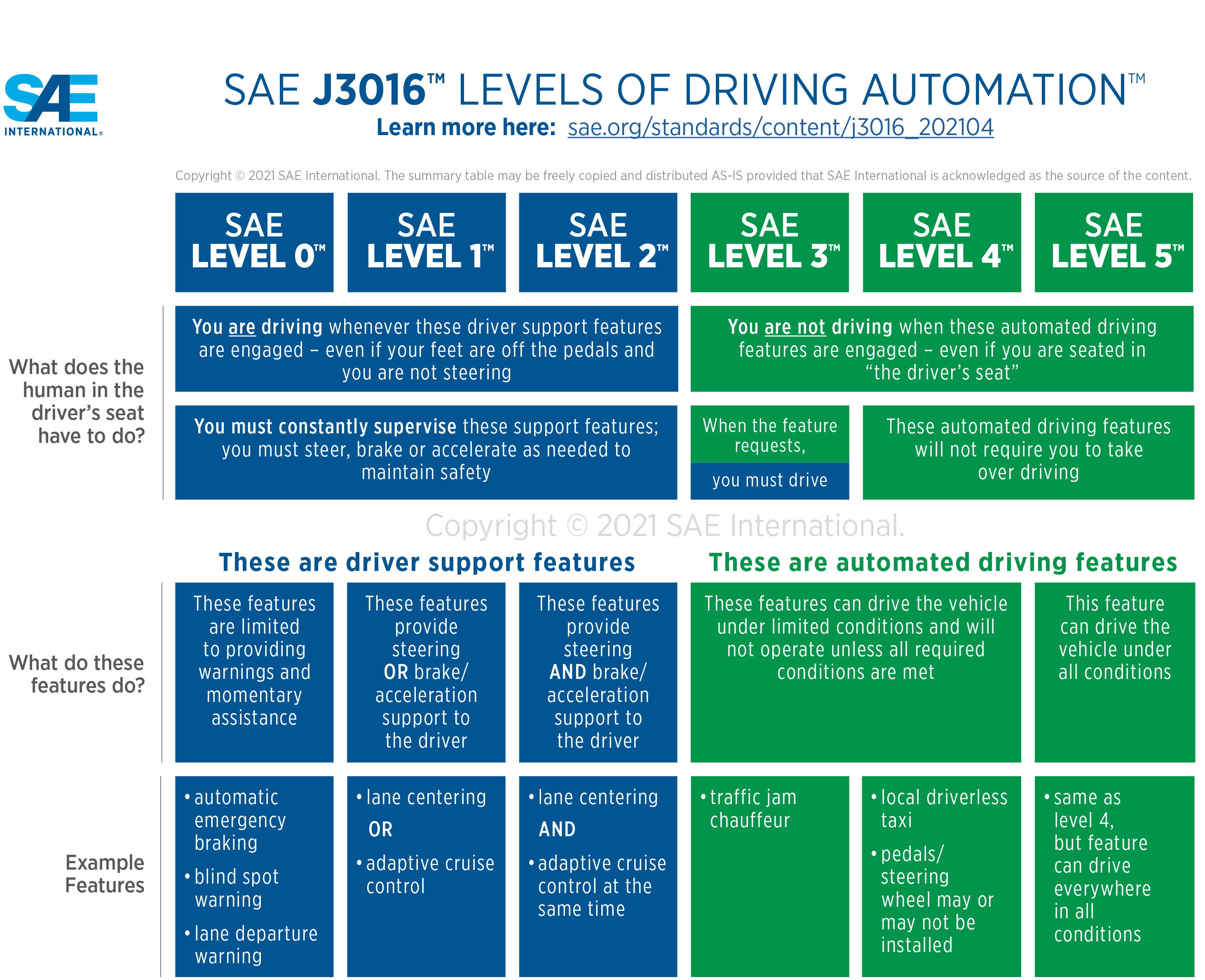}
  \caption{SAE six levels of vehicle autonomy}
  \label{fig:SAE 6 levels}
\end{figure}

According to the Society of Automotive Engineers (SAE) International association, there are six levels of vehicle autonomy based on the level of human and computer involvement \cite{b5}. Level 2 systems include lane centering and adaptive cruise control as driver support features but offer drivers the chance to take control of the vehicle. Although Level 2 systems can be found in many production vehicles, they often have trouble accounting for volatility due to changing weather conditions \cite{b14}. In our research, we attempt to reconcile these difficulties using machine vision systems.

Previous research on Level 2 systems used traditional algorithms without deep learning integration \cite{b12}. The researchers involved in this study were concerned with slow processing speeds and overfitting to training data. However, there have been recent advancements in deep learning for automated vehicles \cite{b2}. In our study, we endeavor to understand how hand-crafted computer vision-based algorithms compare to deep learning algorithms, and whether the two approaches could be combined.

Thus, the goal of this research is to develop, analyze, and evaluate two vehicular lane-keeping algorithms under dynamic weather conditions using real-street electric vehicles. One of these algorithms uses image segmentation and Hough lines, whereas the other algorithm employs an end-to-end convolutional neural network (CNN) with linear regression. We aim to compare the performance of the algorithms to each other, as well as to an ordinary human driver.

\section{Literature Review}

The architecture of self-driving cars is generally split into perception and decision-making \cite{b2}. The perception system estimates how the vehicle relates to its environment, using data from on-board sensors, whereas the decision-making system is responsible for navigation. The localizer subsystem connects these systems, as it estimates the vehicle’s position relative to its surroundings \cite{b2}. Usually, localizers are Global Position System (GPS)-based, Light Detection and Ranging (LIDAR)-based, or camera-based.

Preprocessing methods of lane detection aim to remove irrelevant image parts and enhance the feature of interest \cite{b6}. They use image smoothing techniques, like Median and Gaussian filters, to blur noisy details. At this stage, high dynamic range (HDR) imaging algorithms can remedy the effects of direct sunlight \cite{b10}. The other parts of the preprocessing process are extraction of the region of interest (ROI) and inverse perspective mapping (IPM) to relocate the pixels to a different position \cite{b6}. Various deep learning methods have also been developed for lane following, including CNN- and Long-Short Term Memory (LSTM) \cite{b13}.

On the other hand, vehicle steering methods include spring method center approximation and shifted line following \cite{b7}. Spring center approximation begins with Canny edge detection and Hough line detection. The primary goal of the algorithm is to push the center of the vehicle to the center of the lane with spring physics \cite{b7}. On the other hand, the shifted line following locates the rightmost line in the camera view with blob detection and steers the vehicle at a shifted distance from the line \cite{b7}. Some deep learning algorithms have also been proposed, including end-to-end deep learning \cite{b4}. Through our research study, we aim to use Hough lines, too, but offer a different approach.

Lastly, object detection is of paramount importance, so that a vehicle can follow road signals and avoid collisions. There are many object detection methods using deep learning \cite{b15}. In our research study, we are solely concerned with red line detection to simulate vehicle behavior near a stop sign. With simple cases like these, deep learning methods are often not necessary.

\section{Methods}
We start by explaining our experimental setup, as well as the structure of our software architecture. Then, we discuss the combined image segmentation- and hand-crafted algorithm. Next, we elaborate on our end-to-end deep learning model. Finally, we go over our red detection node.

\subsection{Experimental Setup}
\subsubsection{Simulation}
We use the Robot Operating System (ROS) and Python to develop our algorithms. We use OpenCV, a Computer Vision library, to capture and work with images. We use TensorFlow and Keras for our deep learning. We test the code on simple-sim, a 2-dimensional simulator, as opposed to Gazebo, a 3-dimensional simulator with physics capabilities. Our tests were focused on flat environments, thus the extra computation time of a 3-dimensional sim was not necessary.

\subsubsection{Environment}

\begin{figure}[tb]
    \begin{subfigure}{0.5\columnwidth}
        \centering
        \includegraphics[width=\linewidth]{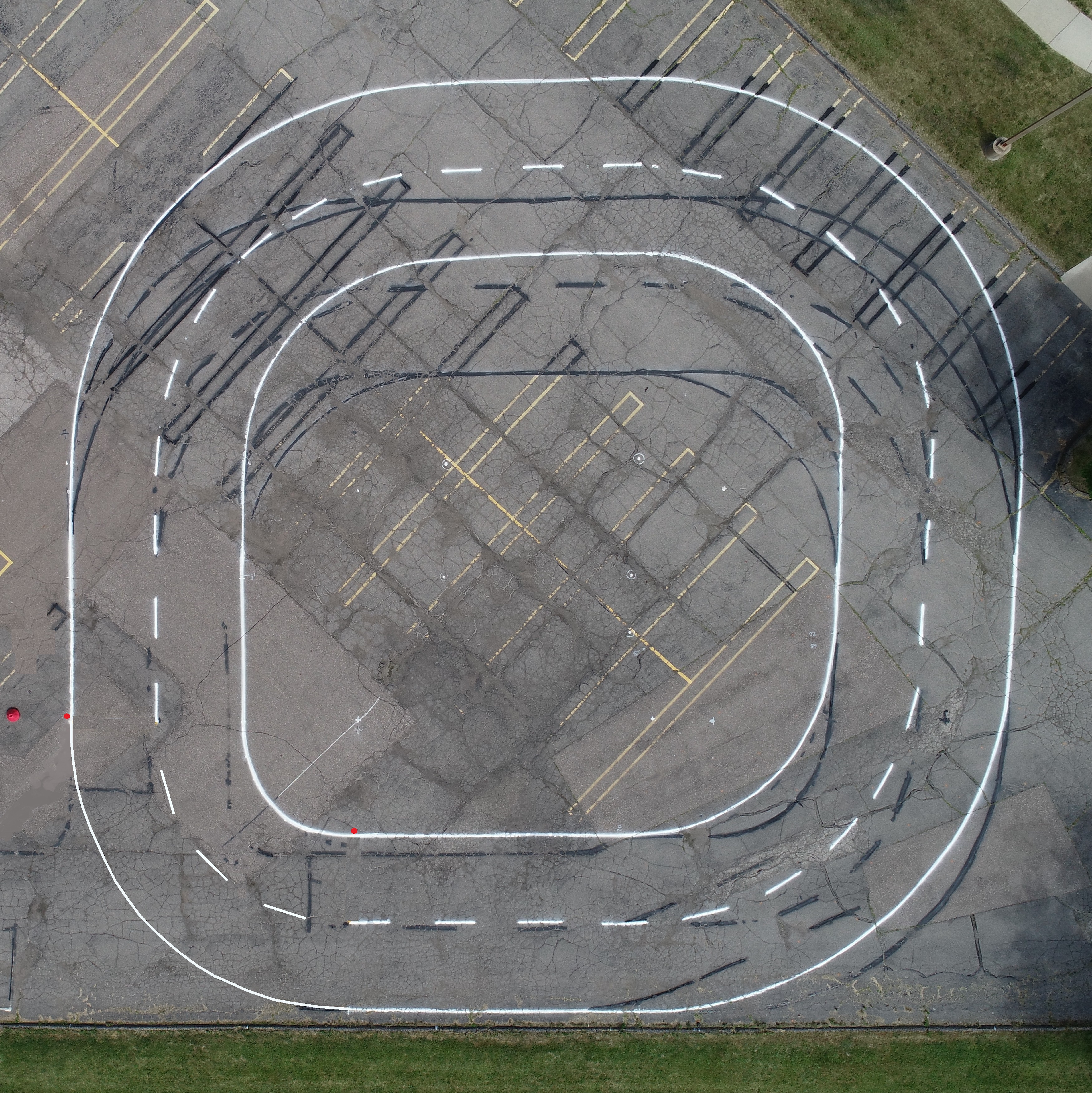}
        \caption{Our test course}
        \label{fig:course}
    \end{subfigure}%
    \begin{subfigure}{0.5\columnwidth}
        \centering
        \includegraphics[width=\linewidth]{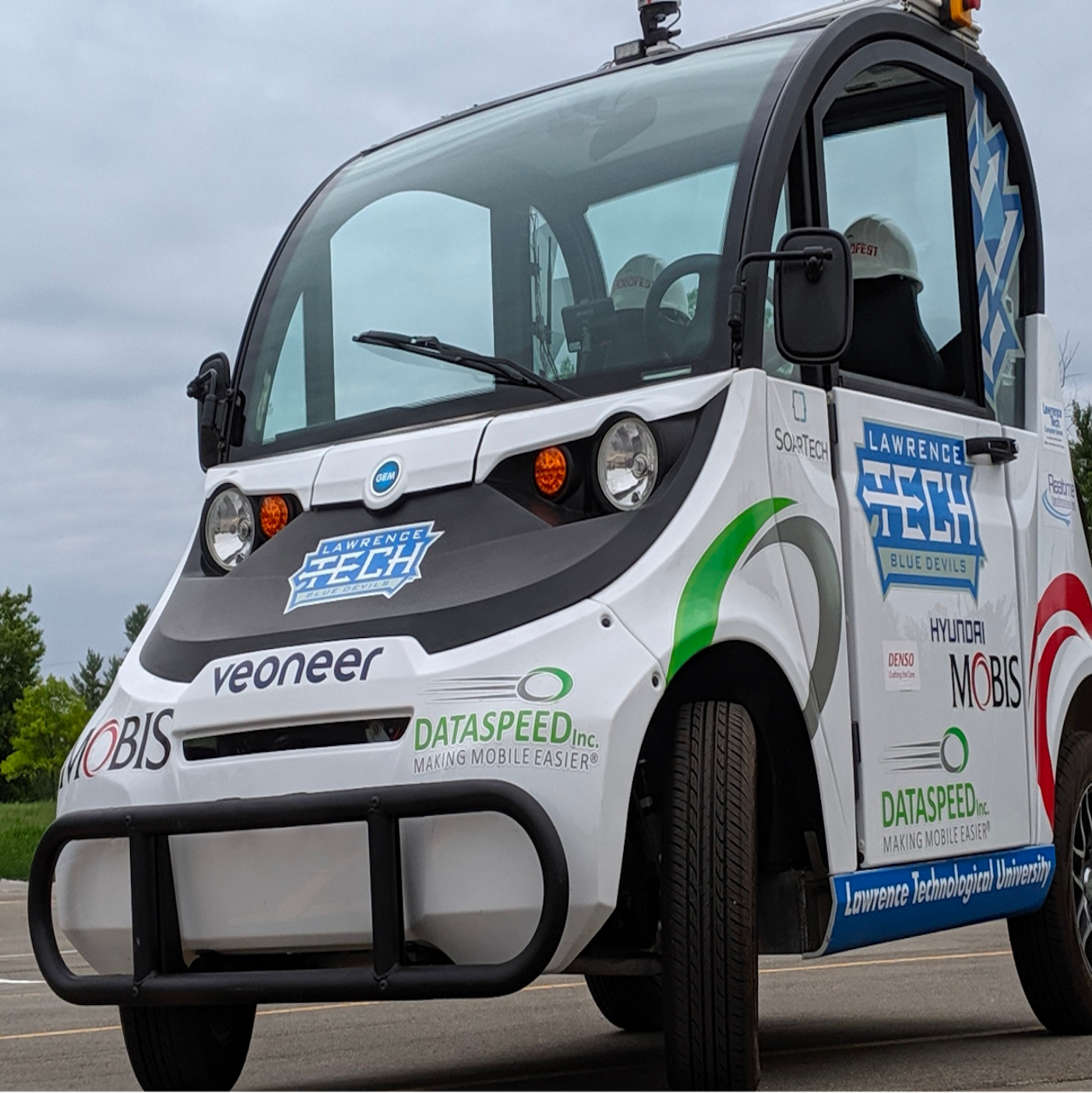}
        \caption{Our vehicle}
        \label{fig:vehicle}
    \end{subfigure}
    \caption{Our real-time experimental setup}
    \label{fig:environment}
\end{figure}

The test course is located in Lawrence Technological University (LTU) Parking Lot H in Southfield, Michigan, a course with rough surfaces, potholes, tight corners, and thin, fading road lines. The brightness and clarity of lines on the course range over different times of day and inclement weather conditions. There are two red markers, one on the outer portion of the outer lane and the other on the inner portion of the inner lane, to mark the start of the course. Both are located over straight portions of the rounded square. The vehicle starts behind either marker and proceeds to drive along the lane until either 5 laps pass or an error occurs.

\subsubsection{Vehicle Specifications}
We use a modified Polaris Gem e2 vehicle, generously sponsored by Mobis and Dataspeed. We call this vehicle Autonomous Campus TranspORt (ACTor) 2. Our vehicle is equipped with a drive-by-wire (DBW) system, vision sensors, 2-dimensional and 3-dimensional Light Detection and Ranging (LIDAR), global positioning system (GPS), and on-board Ubuntu Linux computers. The Polaris Gem e2 has a maximum speed of twenty miles per hour and the battery lasts for twenty miles. We primarily use a forward-facing Mako G-319 Camera with a resolution of 2064 × 1544 pixels, maximum frame rate of 37 frames per second, and native ROS support.

\subsection{Software Architecture}

\begin{figure}[b]
    \centering
    \includegraphics[width=\columnwidth]{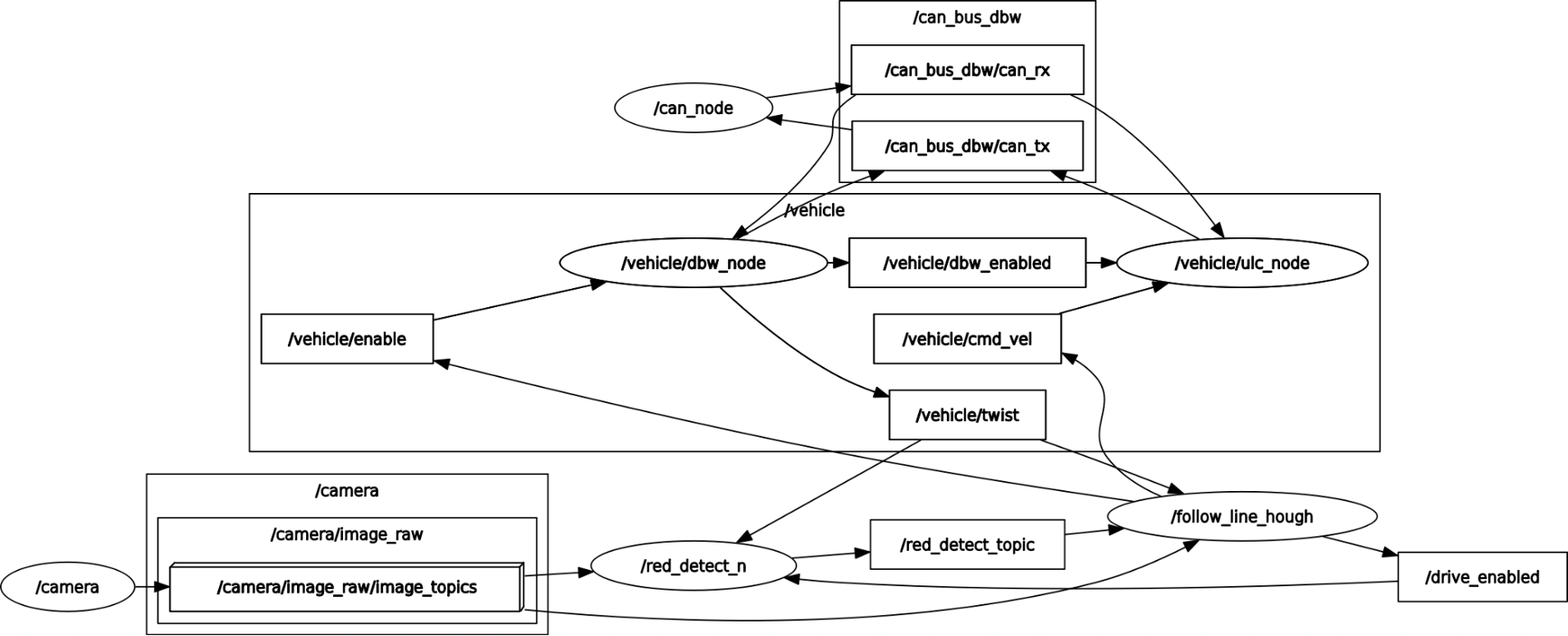}
    \caption{ROS rqt graph for the hybrid model}
    \label{fig:hybrid}
\end{figure}

Our lane-following algorithms are embedded in the Follow Lane Hough and Deep Lane Follow nodes, respectively. They subscribe to the Camera Image Raw topic for continuous Blue-Green-Red (BGR) camera feedback. Our lane-following nodes also subscribe to the Vehicle Twist node to keep track of their forward-linear x and angular z velocities, which we refer to as their speed and yaw rate respectively. They then publish to Vehicle Command Velocity and Drive Enabled topics to engage the DBW system. This is how they drive around the track. Additionally, our Red Detect node subscribes to the Camera Image Raw topic to know when a lap has been completed.

In our combined approach, we use a class-based architecture with a camera callback function. As images are continuously received, they are passed to a methods which handle preprocessing, Hough lines, and driving. However, in our end-to-end deep learning node we do not implement a class as we solely have a short camera callback function. It uses the pretrained deep learning model to steer the vehicle.

\subsection{Combined Approach}
\subsubsection{Image Segmentation}

\begin{figure}[tb]
    \centering
    \includegraphics[width=\columnwidth]{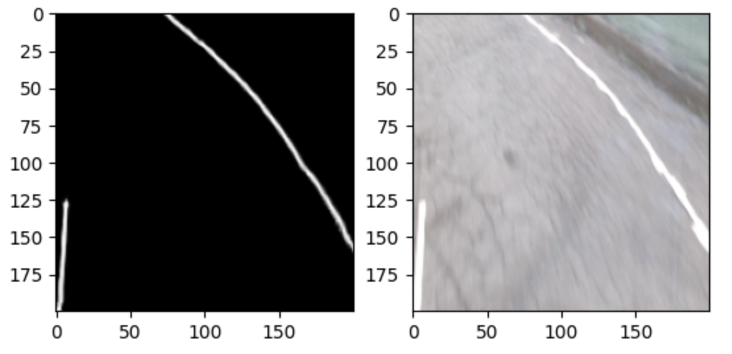}
    \caption{Image segmentation in real-time}
    \label{fig:Segmentation}
\end{figure}
Image segmentation is an image classification task that Artificial Intelligence (AI) models are commonly trained to perform. We decided to use image segmentation to create a simplified image of the road that could be passed into lane following algorithms without concern for noise or many false positives. An AI model trained to perform image segmentation tasks takes an image array as an input. It identifies features it is trained to look for, in our case lines on the road. It then returns an array with the values of the identified objects as 1's, and the rest of the values as 0's. While our segmentation model was only trained to identify the lines on either side of the vehicle, a more robust model could be trained to identify the difference between white and yellow lines, or dashed and solid lines. The biggest drawback when it comes to employing image segmentation is training the AI model. Every image the model is trained on needs to be drawn over by a human. Data augmentation can speed up this process, but it still significantly slows down the training of the model.

\subsubsection{Hand-Crafted Algorithm}

After receiving the processed image, probabilistic hough lines are calculated. We used a function out of the cv2 library that takes in an image and outputs a list of lines on the image in the form of two (x,y) coordinates. We did experiment with the standard hough lines function from cv2 as well, but found that the output lines from the probabalistic function were much easier for the calculations we sought to do and ran faster. Our function is set to filter out lines shorter than 70 pixels, merge gaps between parallel lines of less than 4 pixels, and only have slopes between 0.25 and 100. Each of these values were determined by testing each category manually on our track or in simulation for maximizing true positives while minimizing false positives. The lines each has their center calculated and then are sorted into two lists based on those centers: left and right. We calculate the center points of the left hough lines and right hough lines respectively to project where the lane lines are on each side, and then average these x-values. This new average value is the predicted midpoint of the lane. If the average is farther from the midpoint by 10 pixels, we adjust ACTor 2’s yaw rate by a factor proportional to the difference between the midpoint and center x-coordinate, seen below in Algorithm 1. Together, a hard coded speed, set to 1, and yaw rate are encoded as a Twist message, which is received by the drive-by-wire system at a rate of .34 hertz. Using this Twist message, ACTor 2 drives in the center of the lane.

\begin{algorithm}
    \caption{Steering toward center of Hough lines}
    \SetAlgoLined
    initialization\;
    ratio = self.vel\_msg.angular.z = (mid - center) / mid\;
    \BlankLine
    \uIf{center $<$ mid - 10}{
        self.vel\_msg.angular.z = ratio\;
    }
    \uElseIf{center $>$ mid + 10}{
        self.vel\_msg.angular.z = ratio\;
    }
    \Else{
        self.vel\_msg.angular.z = 0\;
    }
    \BlankLine
    \BlankLine
\end{algorithm}

\begin{figure}[tb]
    \begin{subfigure}{0.5\columnwidth}
        \centering
        \includegraphics[width=\linewidth]{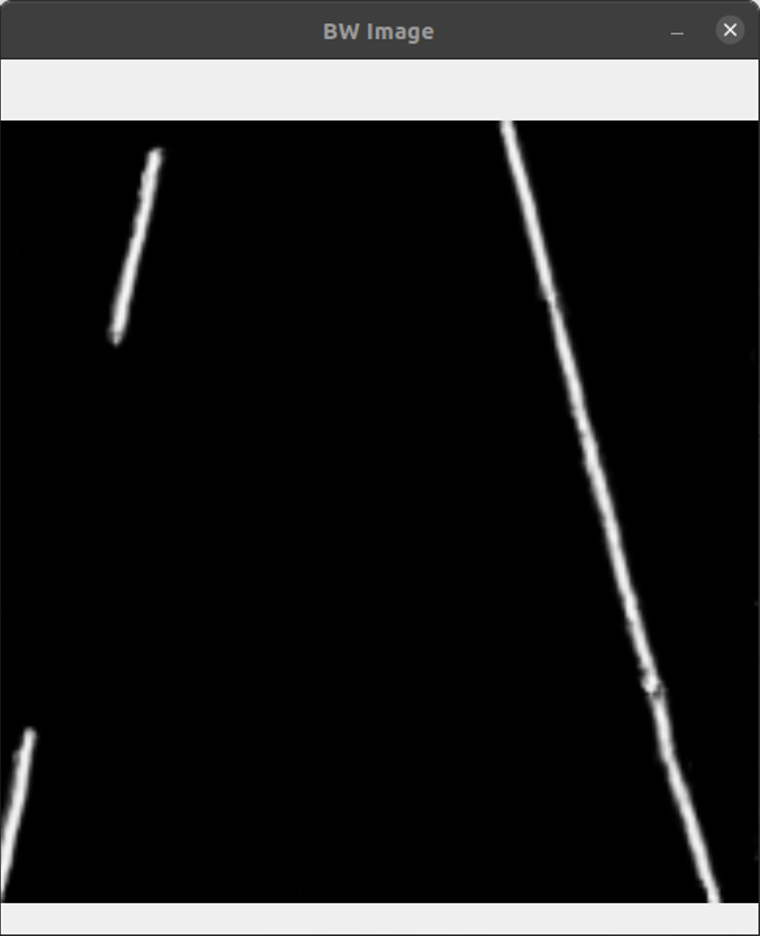}
        \caption{Image segmentation}
        \label{fig:BW Image}
    \end{subfigure}%
    \begin{subfigure}{0.5\columnwidth}
        \centering
        \includegraphics[width=\linewidth]{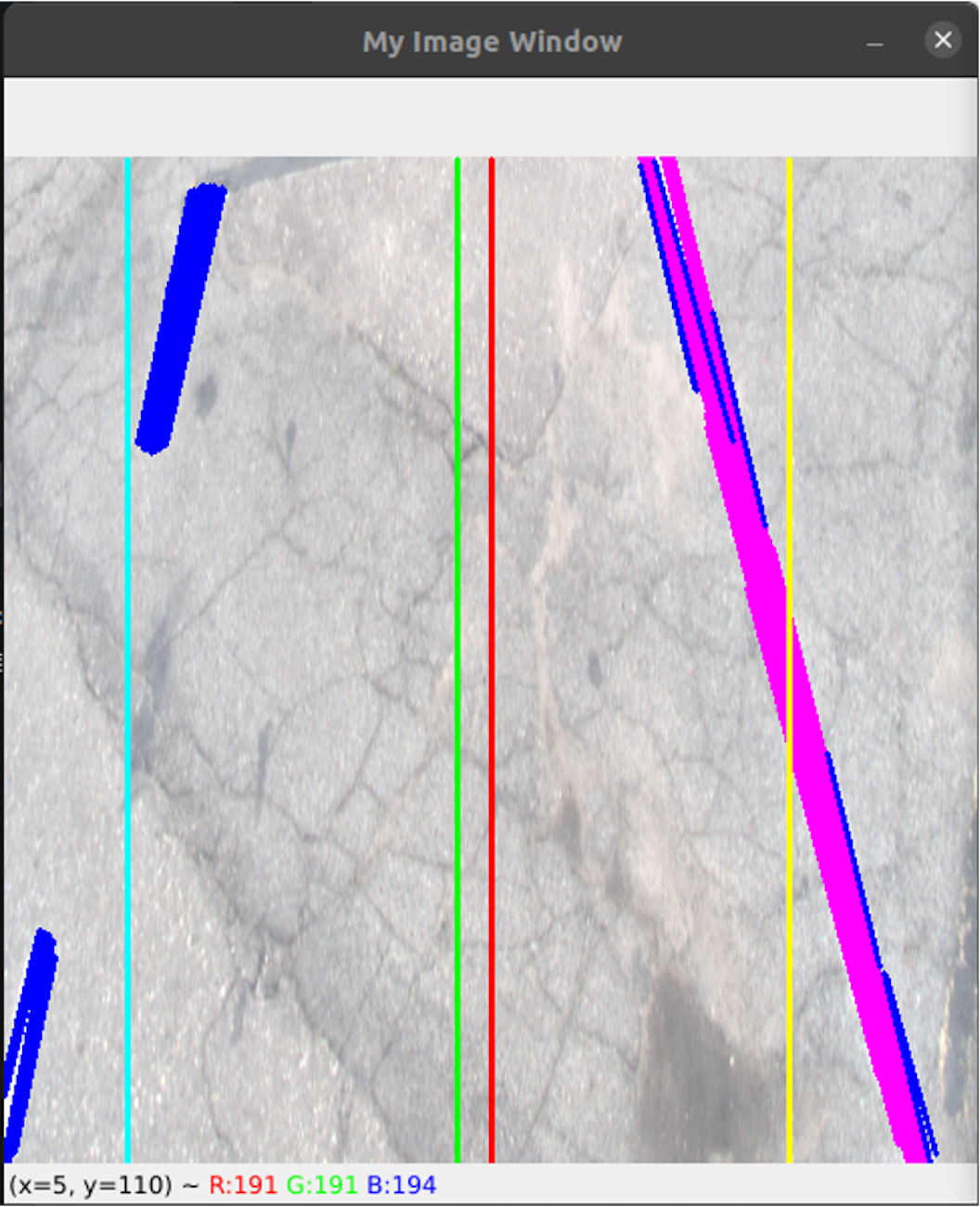}
        \caption{Steering}
        \label{fig:Hybrid Image}
    \end{subfigure}
    \caption{Both parts of our real-time hybrid algorithm}
    \label{fig:environment}
\end{figure}

The cyan line denotes where the algorithm identifies the left lane line, with a similar process for the yellow line and right lane line. Red marks the center of the image and green marks the center of the right and left projected lanes. The blue and magenta lines represent identified hough lines, effectively where the car sees the lane lines. Magenta denote lines longer than 180 pixels. Shown in Figure 5b, the car would turn slightly left based on the information given.

\subsection{End-to-End Deep Learning}
\subsubsection{Training} Implementing deep learning algorithms requires training the models through a variety of scenarios to ensure consistent performance under any conditions. To train our end-to-end deep learning model, we created a program to automatically pair the images captured from our camera with the steering angle of the vehicle. This data could quickly be turned into a database that can be used to train a TensorFlow model. 
\subsubsection{Data Augmentation}We made sure to gather data in different light and weather conditions, but we could not gather enough data just by driving, so we implemented data augmentation. Data augmentation is commonly used to train deep learning models; it involves slightly altering the inputs and pairing them up with the same outputs. This teaches the model to ignore slight differences in data, making the model more consistent. In our case, we augmented the data by adjusting the brightness, contrast, orientation, and color map of the input images. For every 1 input image, we augmented 9 more inputs, augmenting our training data by a factor of 10.
\begin{figure}[tb]
    \centering
    \includegraphics[width=\columnwidth]{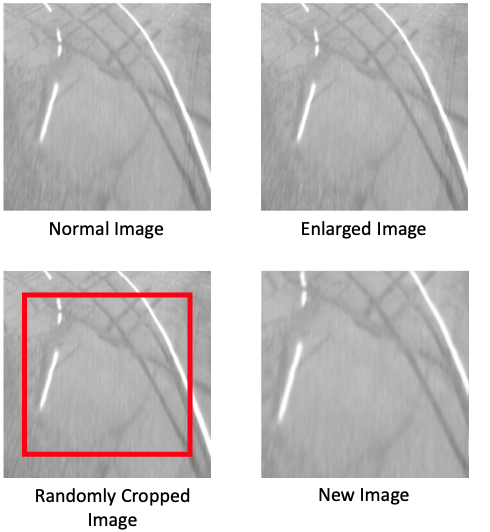}
    \caption{Example of data augmentation}
    \label{fig:Data augmentation}
\end{figure}
\subsubsection{Implementation} When implementing the model, we used common sense measures to ensure an even more consistent result. These measures included limiting how far the model was allowed to turn the wheel every time the script ran. Because we trained the model on modified images, we concluded that the model would perform better if it was allowed ``multiple guesses'' on modified images. For every frame captured by the camera, we created 2 slightly altered images, adjusting the brightness and contrast of the original. Then we fed the original image and the two modified images to the model. Next, we took the 3 outputs from the model, removed outliers that were far off from the previous steering angle, and set the new steering angle to the average of those outputs. This gave the model room for error and ensured that the model would work under even more conditions.

\section{Discussion}
With computational models like ours, it is important to make assumptions. Some assumptions we made were that vehicles did not have access to a Roadside Unit and Wireless Fidelity (Wi-Fi) network. Hence, they could not collaborate with other vehicles using an Intelligent Transportation System \cite{b3}. Another assumption we made was vehicles driving based on input from a sensor, specifically a camera. We restricted ourselves to a single, forward-facing camera in order to focus in on lane-following, as if a car can follow a lane with one camera, other cameras can be focused on other features such as safety. Hypothetically, all the roadways could be navigated using GPS and path planning alongside the camera, but that was not the goal of this study.

\subsection{Failed Approaches}
Although our lane-following algorithms were mostly successful, we experimented with other approaches which worked poorly.

\subsubsection{Dynamic Threshold}
We did not always use image segmentation during our preprocessing step in the hybrid algorithm. At first, we found the rightmost contour in each BGR image (the right lane line). Using OpenCV in Python, we median blurred the images and converted them to grayscale. Next, we chose an arbitrary threshold value to convert the images to black-white (BW). We selected minimum and maximum percentages of white pixels allowed per image and checked the percentage in our current BW image. If this percentage was not within bounds, we adjusted the threshold using binary search until a suitable threshold was found. Then, we drove the vehicle at a constant displacement from the right lane line.

However, our dynamic thresholding did not hold up in adverse weather and volatile lighting conditions. We subsequently pivoted our preprocessing design to use deep learning based on various images.

\subsubsection{DeepLSD}
We experimented with a line segment detection and refinement method leveraging deep image gradients, coined DeepLSD\cite{b11}. We integrated it into our simple-sim simulation, where it detected many of the black road markings as lines, while not marking the white road lines. After early failures, we decided to solely focus on developing the hybrid and end to end deep learning algorithms for the sake of time. Although we did not succeed in implementing a working adaptation of DeepLSD, we would like to further explore its viability in the future.

\subsubsection{Slope-Based Algorithm}
In our hand-crafted steering algorithm, we we were tasked with discovering creative approaches to make ACTor 2 drive in the center of the lane. Given images of the course, we originally postulated that the mean slope of the lane lines on either side of ACTor 2 is related to its yaw rate. However, when we found no relationship between these values, we tried calculating the mean slopes differently. We first weighted the positive and negative slopes equally. Next, we weighted the slopes on the left and right side of the screen equally. We even tried using the median mean slope weighted by the length of the lane lines from the past 10 images to fit a logistic-sigmoid \eqref{eq:Sigmoid} curve with SciPy. Lastly, we tried fitting an arctangent \eqref{eq:Arctangent} curve and piecewise function \eqref{eq:Piecewise}. All of our approaches were unsuccessful.

\begin{align}
    &\frac{L}{1+e^{-k\cdot (x-x_0)}} \label{eq:Sigmoid} \\
    &k\cdot \tan^{-1}(w\cdot (x-x_0)) + y_0 \label{eq:Arctangent} \\
    &\begin{cases}
        \frac{-1}{x+x_0}+k_1, & \text{if } x < r_1 \\
        mx + b, & \text{if } r_1 < x < r_2 \\
        \frac{-1}{x+x_1}+k_2, & \text{if } x > r_2
    \end{cases} \label{eq:Piecewise}
\end{align}
\\

Figure 7 shows a scatter plot of the mean slope of the lane lines on either side of ACTor 2 relative to its yaw rate. To produce this data, we manually drove around the track for 2 laps. The orange line is the baseline model, which is just the arithmetic average of all the slopes. The red line is a linear regression using an 80\%-20\% training-testing dataset split with Scikit-learn.

We ultimately used a different approach to calculate the vehicle steering, as described in our methods. We did not rely on the slope or length of the lane lines, opting instead to direct the vehicle toward the center of the lines.

\begin{figure}[tb]
    \centering
    \includegraphics[width=\columnwidth]{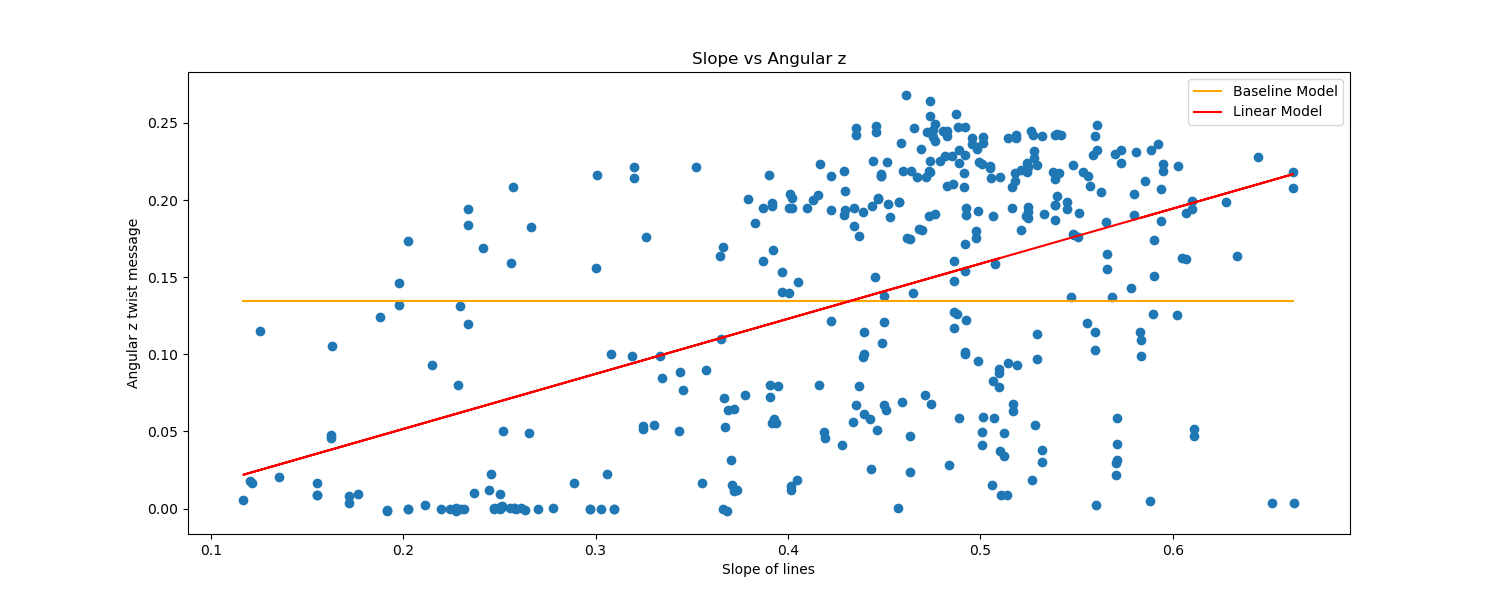}
    \caption{A scatter plot of mean slope vs yaw rate}
    \label{fig:Failed}
\end{figure}

\subsection{Limitations and Future Work}
\subsubsection{Limitations}
We did not evaluate night compatibility for our models. But our camera cannot display images well in the absence of light. To drive the vehicle at night, we could experiment with higher-definition cameras that have night-vision or LIDAR sensors.

One major limitation was all our testing taking place on LTU's Parking Lot H. We are not sure how our methods would fare in other environments. We chose a test course with shadows, jagged surfaces, and sharp corners. We explicitly chose not to drive on other roads near LTU because Level 2 algorithms already work well on straightaways. On the flip side, roads in local neighborhoods pose other challenges like dead reckoning curves, where there are no lines, and parked cars and obstacles.

Another limitation was hardware limitations. We ran our programs every 20 milliseconds, but could not run our programs much faster with the resources given to us.

Moreover, we did not consider robustness to adverserial attacks. In a future study, we can look into security considerations.

\subsubsection{Future Work}
In the future, we would like our Follow Lane Hough and Deep Lane Follow nodes to run simultaneously. Currently, they publish to the Vehicle Command Velocity and Drive Enabled Topics to deterministically steer the vehicle around the course. However, we would like both nodes to publish perspective speed and yaw rate messages to a Control node, as well as their confidence. The Control node would dynamically weight the outputs of each model to derive new values for the speed and yaw rate. Then, it would steer the vehicle based on these values.

One avenue for further exploration is other partial deep learning approaches. For instance, we can use Recurrent CNN's to label environments other than our test course \cite{b9}. Scene classification like \cite{b9} is a good way to extract features of representative objects. Then, values for parameters in our models would be updated accordingly. Furthermore, an alternative approach to image segmentation is using unsupervised training.

Other ideas we have are end-to-end deep learning approaches that are not based on linear regression. There is not much literature on reinforcement learning for self-driving vehicles.

\subsection{Conclusion}
We presented two vehicular lane-following algorithms. The first algorithm uses a combined traditional- and deep-learning approach, whereas the second algorithm uses end-to-end linear regression-based deep learning. The hybrid algorithm completes 5 laps on the inner and outer lane, whereas the second algorithm fails on the inner lane. In the future, we need to continue working on Level 2 algorithms in dynamic weather conditions, and ultimately implement these algorithms in production vehicles. Self-driving algorithms of this type are becoming increasingly important, as they can reduce vehicular fatalities on the road. In the robotics, autonomous vehicles, and computer science fields, it is imperative that we collaborate to increase safety on the roadways by further designing lane-following algorithms.

\section*{Acknowledgment}

Our work was supported by the National Science Foundation under Grants No. 2150292 and 2150096. We thank our mentors, Joe DeRose and Nick Paul, as well as our teaching assistants, Ryan Kaddis and Justin Dombecki.

\end{document}